# MOTION PLANNING ON AN ASTEROID SURFACE WITH IRREGULAR GRAVITY FIELDS

Himangshu Kalita,[*] and Jekan Thangavelautham,[†]

There are thousands of asteroids in near-Earth space and millions in the Main Belt. They are diverse in physical properties and composition and are time capsules of the early solar system. This makes them strategic locations for planetary science, resource mining, planetary defense/security and as interplanetary depots and communication relays. However, asteroids are a challenging target for surface exploration due it its low but highly nonlinear gravity field. In such conditions, mobility through ballistic hopping possess multiple advantages over conventional mobility solutions and as such hopping robots have emerged as a promising platform for future exploration of asteroids and comets. They can traverse large distances over rough terrain with the expenditure of minimum energy. In this paper we present ballistic hopping dynamics and its motion planning on an asteroid surface with highly nonlinear gravity fields. We do it by solving Lambert's orbital boundary value problem in irregular gravity fields by a shooting method to find the initial velocity required to intercept a target. We then present methods to localize the hopping robot using pose estimation by successive scan matching with a 3D laser scanner. Using the above results, we provide methods for motion planning on the asteroid surface over long distances. The robot will require to perform multiple hops to reach a desired goal from its initial position while avoiding obstacles. The study is then be extended to find optimal trajectories to reach a desired goal by visiting multiple waypoints.

## INTRODUCTION

The millions of asteroids in the solar system are known to be diverse in size, shape and composition. Exploration of asteroids can give further insight into the origin questions, namely the origin of the solar system, origin of Earth and origin of life. Certain C-type asteroids are known to contain water-rich carbon based organic-molecules and rare-minerals. Some of these asteroids are hypothesized to have seeded Earth with the needed organic building-block to start life. Some asteroids contain remains of existing planets and serve as time-capsules that can provide pristine records of the early geology and geohistory of the planets. Some of these asteroids are known to be composed of material billions of years old. These small bodies are remnants of planet formation, progenitors of meteorites, and are therefore high-value targets in the Planetary Science Decadal Survey. The exploration of these small-bodies can give us insight into the formation of the solar-system, planetary defense and future prospect for in-situ resource utilization. Exploration of these asteroid can also provide useful insight into the primordial solar system.

---

[*] PhD Student, Aerospace and Mechanical Engineering, University of Arizona, US, 85721.
[†] Assistant Professor, Aerospace and Mechanical Engineering, University of Arizona, US, 85721.

The exact composition of these asteroids is hard to discern, as they have undergone layering and surface weathering processes that modify chemical composition of the surface. Flyby and long-range observation of asteroids is insufficient to determine what is beneath the top layer of the asteroid. Only surface and subsurface exploration of asteroids can answer these questions. The recent exploration of asteroids such as Bennu and Ryugu all show evidence for the 'brazil nut' effect, where large objects end up on the surface compared to small objects. This makes surface extremely rugged and extremely challenging for surface exploration. This also further confirms that the material underneath will be substantially different. All of these factors reinforce the need for surface exploration and use of in-situ instruments to analyze composition of the surface and subsurface material. We have identified the need to perform excavation to perform chemical analysis, penetrometery and seismic analysis as high-priority science.[1]

However, the surface environment of asteroids presents many unique challenges and opportunities. Unlike Earth, Mars or Moon, asteroid gravity is low (10-1,000 μg) and their irregular shape results in highly irregular gravity fields. This eliminates the practical use of conventional multi-wheeled rovers that rely on surface traction. Instead, hopping rovers are more naturally suited for such environments, as they can traverse large distances over arbitrarily rough terrain with the expenditure of little energy and little need for traction. Although the cost of hopping over an asteroid surface is significantly low, too high a thrust can result in a rover attaining escape velocity and tumbling into space. Thus, the surface operation will be limited by the local gravitational field.

As communication suffers from large delays over very long distances, the hopping rover operating on an asteroid surface will require a high degree of autonomy in addition to controllability. The rovers deployed on asteroid surfaces to date relies on a mothership relay, however they are infrequent. Wheeled rovers operate through continuous interaction with the environment, and as such path planning is performed through visual perception and terrain classification. However, hopping rovers can only apply forces from rest on the surface and have no control of their trajectory mid-flight. Thus, a sequential architecture for hopping rover autonomy is required that can plan ahead before executing any hop. Mothership-independent localization feedback is also required for the hoping rovers to be autonomous as interaction with the mothership is infrequent.

In this paper we present a dynamics model for a rover deployed on the surface of an asteroid followed by ballistic hopping simulations and a self-localization method by using a 3D LiDAR sensor onboard the rover. Ballistic hopping and self-localization methods are then used to develop an algorithm that provides optimal trajectories for a rover to reach a target position from its initial position. The algorithm can also be used to guide the rover through multiple waypoints in between the initial and target position. Finally, an algorithm is demonstrated for multiple rovers to parallelly explore maximum area on an asteroid surface while maintaining a desired communication link among them. These very same tactics to perform asteroid surface and subsurface science is needed for resource prospecting and mining. Asteroids are rich in resources required for a space economy, including sources of propellent, structural materials and rare minerals. The low gravity of asteroids makes it especially appealing for supplying resources in cis-lunar and interplanetary space. There are also many asteroids in strategic locations to be future communication relays,[2] propellant sources and space construction material.

In the following section, we present background followed by the dynamic model of the rover and asteroid environment. Next, we present the dynamics of ballistic hopping, localization and motion planning to enable a hopping rover to traverse an asteroid followed by conclusions and future work.

## BACKGROUND

Much work has been done in observing asteroids by ground-based telescopes and space observatories but exploring asteroid surfaces with landers is a major challenge. Several asteroid sample return missions have been launched and several others are being studied worldwide. Japan Aerospace Exploration Agency (JAXA) developed an unmanned spacecraft named Hayabusa to return a sample of material from a small near-Earth asteroid named 25143 Itokawa to Earth for further analysis. Hayabusa studied the asteroid's shape, spin, topography, color, composition, density and history and finally landed in November 2005 and collected tiny grains of asteroid material.[3] The spacecraft also carried a 591g small rover named MINERVA (Micro/Nano Experimental Robot Vehicle for Asteroid) which unfortunately hopped off the surface of the asteroid and tumbled into space.[4] The lessons learned from the successful Hayabusa I mission led to the development of Hayabusa II asteroid sample return mission. Hayabusa II was launched in December 2014 and reached its target asteroid, 162173 Ryugu (1999 JU3) in July 2018. The Institute of Space Systems of the German Aerospace Centre (DLR) in cooperation with French space agency (CNES) built a small lander called MASCOT (Mobile Asteroid Surface Scout) to complement the sample return mission. MASCOT carries an infrared spectrometer, a magnetometer, a radiometer and a camera, and can lift off the asteroid to reposition itself for further measurement.[5] Both MINERVA and MASCOT are equipped with internal momentum devices for hopping with minimal control.

Another asteroid sample return mission is NASA's Origins, Spectral Interpretation, Resource Identification, Security, Regolith Explorer (OSIRIS-REx) mission led by University of Arizona and was launched in September 2016. It reached its target asteroid 101955 Bennu in December 2018. The spacecraft will measure Bennu's physical, geological, and chemical properties and collect at least 60 g of regolith.[6] Rosetta is another spacecraft built by the European Space Agency, launched in March 2004 and performed detailed study of comet 67P/Churyumov-Gerasimeko. Rosetta carried a ~98 kg lander named Philae which performed studies on elemental, isotopic, molecular and mineralogical composition of the comet, characterized the physical properties of the surface and subsurface material and the magnetic and plasma environment of the nucleus. However, most of these missions perform fly-bys, and touch and go operations to mitigate the risks of 'landing' on an asteroid.

Work has also been done in developing hopping mechanisms for low-gravity environments. A typical approach to hopping is to use a hopping spring mechanism to overcome large obstacles.[7] One is the Micro-hopper for Mars exploration developed by the Canadian Space Agency.[8] Another technique for hopping developed at MIT utilize Polymer Actuator Membranes (PAM) to load a spring.[9,10] Other techniques for hopping mimic the grasshopper and use planetary gears within the hopping mechanism. NASA JPL and Stanford also developed a planetary mobility platform called "spacecraft/rover hybrid" dubbed "Hedgehog" that relies on internal actuation through three mutually orthogonal flywheels and external spikes.[11] Another autonomous microscale surface lander developed is PANIC (Pico Autonomous Near-Earth Asteroid In-Situ Characterizer) that utilizes hopping as a locomotive mechanism in microgravity.[12]

In our earlier work, we have proposed Sphere-x[13,22] and AMIGO[14] that uses chemical and sublimate-based propulsion to perform hopping on low gravity environments. With so much work done on these hopping rovers, autonomous architecture for path planning in irregular gravity fields has a wide scope of research. Analytical hopping control laws has been derived on a smooth, spherical[15] and ellipsoidal[16] asteroid model. But these studies assumed a highly simplified dynamics model. Another method is to use a data-driven approach, using high-fidelity dynamics model and various uncertainty models.[17] Our approach also uses a high-fidelity dynamics model to perform trial and error evaluation in simulation of candidate maneuvers to determine their suitability. The trial and error evaluation take into account risk, rewards and local uncertainties. Using this method, it is then possible to execute optimal or near-optimal hopping maneuvers to get from one location to another on the asteroid surface.

**DYNAMICS MODEL**

The asteroid body is represented as a closed polygonal mesh with $k$ triangular facets, where each facet $F_i$ has outward normal $N_i$. Figure 1 shows the coordinate system used throughout the paper. $\hat{I} = \{\hat{I}_x \quad \hat{I}_y \quad \hat{I}_z\}$ is the inertial frame and $\hat{b} = \{\hat{b}_x \quad \hat{b}_y \quad \hat{b}_z\}$ is the asteroid's body fixed frame, with the origin coinciding with the center of mass of the asteroid. The asteroid rotates at a constant angular velocity $\omega = \Omega \hat{b}_z$. A rover is represented by its position vector $r$ and velocity vector $v$ relative to the asteroid body frame $\hat{b}$.

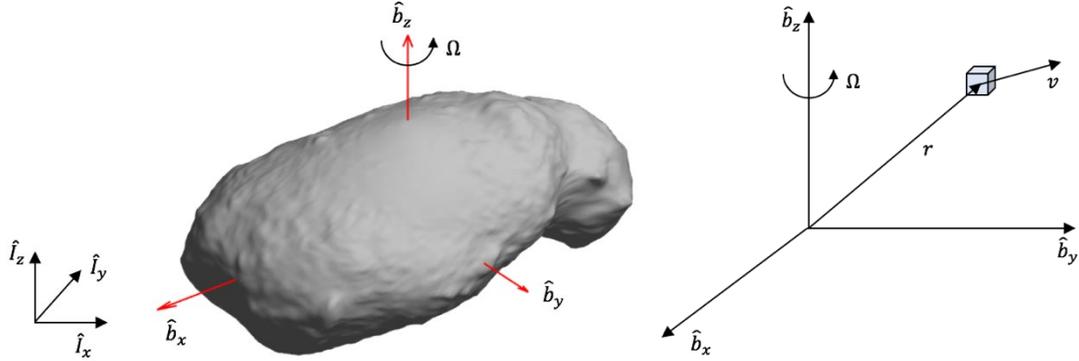

**Figure 1: (Right) Inertial frame $\hat{I}$ and body fixed frame $\hat{b}$ defined for the asteroid Itokawa. (Right) Position and velocity vector of the rover defined with respect to the body fixed frame $\hat{b}$.**

The dynamic equation of motion of the rover in the asteroid's body fixed coordinate system is expressed as Eq. (1).

$$\ddot{r} + 2\omega \times \dot{r} + \omega \times (\omega \times r) + \dot{\omega} \times r = g + d + u \qquad (1)$$

where, $g$ is the gravitational acceleration, $d$ is the disturbance acceleration such as SRP and third body perturbations, and $u$ is the control acceleration. Also, the asteroid is considered to have a fixed angular velocity, so $\dot{\omega}$ is equal to zero. Although gravity in smaller bodies are weaker than on Earth, it still is the dominant force on rovers. The polyhedral model is the most accurate gravity model for smaller irregular bodies which leverages the divergence theorem to exactly model the gravitational potential ($U$), gravitational acceleration ($g = \nabla U$), gradient ($\nabla \nabla U$) and Laplacian ($\nabla^2 U$) of a constant density polyhedron as a summation over all facets and edges of the surface mesh as shown in Eq. (2).[18]

$$U(r) = -\frac{1}{2}G\rho \sum_{e \in edge} r_e^T E_e r_e \cdot L_e + \frac{1}{2}G\rho \sum_{f \in face} r_f^T F_f r_f \cdot \Theta_f \qquad (2)$$

Where, $r_e$ is a vector from the field point to an arbitrary point on each edge, $E_e$ is a dyad defined in terms of the face and edge normal vectors associated with each edge, $L_e$ is a logarithmic term expressing the potential of a 1D straight wire, $r_f$ is a vector from the field point to an arbitrary point on each face, $F_f$ is the outer product of face normal vectors, and $\Theta_f$ is the solid angle subtended by a face when viewed from the field point.

## BALLISTIC HOPPING

The rover needs to hop from rest at position $r_0$ with velocity $v_0$ and impact at position $r_f$ with velocity $v_f$. The problem of computing the launch velocity, $v_0$ to intercept a target location, $r_f$ at time $\tau = t_f - t_0$ is the well-known "Lambert orbital boundary-value problem" and efficient numerical solutions for different types of gravity fields are available. For the case of asteroids with irregular gravity field, a simple shooting method is used to calculate the launch velocity to successfully impact a target location. First an initial guess of the initial velocity $v_{0(1)}$ is created by solving the two-body form of Lambert's problem for a uniform spherical gravity field. At each shooting iteration $i$, the initial delta-velocity vector is updated according to Eq. (3).

$$\Delta v = [\Phi]^{-1} \Delta r \tag{3}$$

$$\Phi = \left[\frac{\partial r}{\partial v_0}\right] = \begin{bmatrix} \partial x/\partial \dot{x}_0 & \partial x/\partial \dot{y}_0 & \partial x/\partial \dot{z}_0 \\ \partial y/\partial \dot{x}_0 & \partial y/\partial \dot{y}_0 & \partial y/\partial \dot{z}_0 \\ \partial z/\partial \dot{x}_0 & \partial z/\partial \dot{y}_0 & \partial z/\partial \dot{z}_0 \end{bmatrix} \tag{4}$$

Where, $\Phi$ is the state transition matrix, the error in the final position vector $\Delta r$ is determined from the difference between the desired final position, $r_f$ and the final position vector predicted by numerical integration of the equations of motion of the rover, $r_n$ as shown in Eq. (5)

$$\Delta r = r_f - r_n \tag{5}$$

The new initial vector is then updated as Eq. (6).

$$v_{0(i+1)} = v_{0(i)} + \Delta v \tag{6}$$

Figure 2 shows multiple hopping trajectories for a rover from its initial position of $r_0 = [-183.3 \quad 45.81 \quad 70.15]m$ to its target position of $r_f = [85.58 \quad -86.16 \quad 8.15]m$ for transfer times $\tau = 0.5hr, 0.75hr, 1hr, 1.25hr, 1.5hr$.

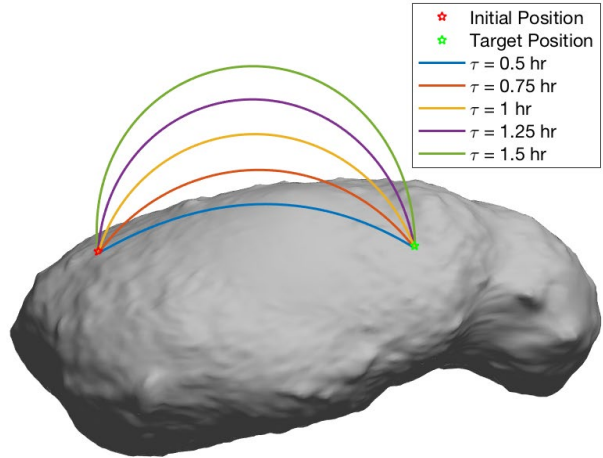

**Figure 2: Hopping trajectories from an initial position to its target position for different transfer time.**

## LOCALIZATION

The rover is assumed to be equipped with a 3D LiDAR sensor which can acquire 3D point cloud of the surrounding environment at a particular frequency. Two successive point clouds are matched to perform self-localization of the rover. The Iterative Closest Point algorithm is used in this paper which tries to minimize the Euclidean distance between an input data and a reference model to find the transformation between the two sets of data.[19] Consider two sets of 3D points, model set $M$ (with $n_m$ points) and data set $D$ (with $n_d$ points) $\subseteq \mathbb{R}^3$. The objective is to find a transformation function $u: D \to M$ that minimizes the mean squared distances ($D_{ms}$) between $D$ and $M$ as shown in Eq. (7)

$$D_{ms}(D, M, u) = \frac{1}{n} \sum_{m \in M, d \in D} \|m - u(d)\|^2 \qquad (7)$$

Incorporating the rotation ($R$) and translation ($t$) matrices into the matching function, the minimization problem can be written as Eq. (8)

$$\min_{u:D \to M} \frac{1}{n} \sum_{i=1}^{n} \|m_i - Rd_i - t\|^2 \qquad (8)$$

With this mathematical formulation, the ICP algorithm tries in each iteration to minimize the $D_{ms}(D, M, u)$ by switching between a matching and a transformation stage. In the matching stage, the objective is to minimize the mean squared distances $D_{ms}(D, M, u)$ by finding the best correspondence between a point $d_i \in D$ and $m_i \in M$. During the transformation stage, the objective is to compute the optimal $R$ and $t$ that minimizes (5). The rotation matrix $R$ and the translation matrix $t$ are then used to find the orientation and position of the rover with respect to its initial position. Figure 3 (left) shows the actual and estimated trajectory of a rover by performing scan matching of 3D point clouds at a frequency of 0.5 Hz. Figure 3 (right) shows the error between the actual trajectory and the estimated trajectory. It can be seen that the error is less than 1 m for a hopping distance of ~110m. The error can be minimized further by registering 3D point cloud samples at a higher frequency.

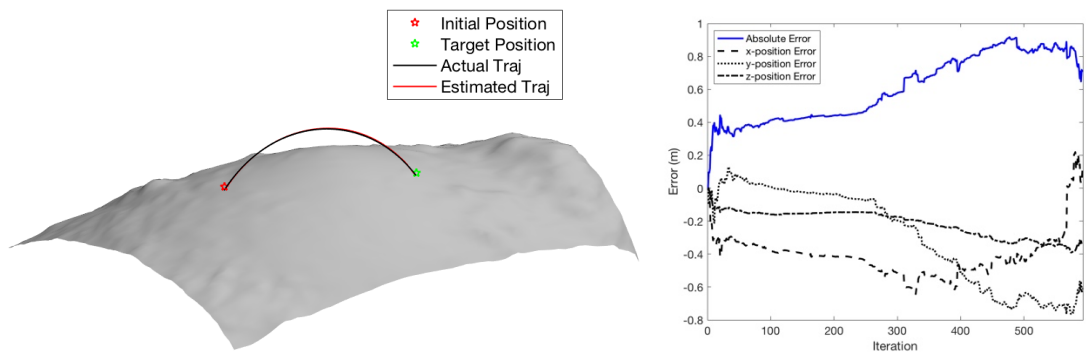

**Figure 3: (Left) Actual trajectory and the estimated trajectory by 3D point cloud scan matching for a single hop. (Right) Absolute error and errors along x, y and z between the actual trajectory and the estimated trajectory.**

MOTION PLANNING

For traversing long distances, the rover needs to plan and perform multiple sequential hops to reach a desired goal from its initial position while avoiding collision. We developed a probabilistic path planner using random sampling which is then optimized using evolutionary algorithms. The planner builds a path from the start position to the goal position by randomly sampling points in the search space and linking them with admissible constrained hopping trajectories as shown in Figure 4 (left). The search space for the random sampler is the polygonal mesh $P$ of the asteroid and the random sampler needs to search a continuous path from an initial position $r_{init} \in P$ to a goal position $r_{goal} \in P$. A Rapidly-exploring Random Tree, $\mathcal{T}$ is constructed so that all the vertices of the tree are in $P$.[20] The first vertex of $\mathcal{T}$ is $r_{init} \in P$. In each iteration, a random state, $r_{rand}$ is selected from $P$. Step 4 finds the nearest vertex to $r_{rand}$. Step 5 finds a new position $r_{new}$ that steers the rover from $r_{near}$ towards $r_{rand}$ satisfying the maximum hopping constraint of $\Delta$. The new state $r_{new}$ is added as a vertex to $\mathcal{T}$ and a hopping trajectory from $r_{near}$ to $r_{new}$ defined by transfer time $\tau \sim \mathcal{N}(\mu_\tau, \sigma_\tau^2)$ is added as an edge to $\mathcal{T}$. Finally, the algorithm explores $\mathcal{T}$ to return the shortest path connecting $r_{init}$ to $r_{goal}$. Figure 4 (right) shows a random sample created on the surface of asteroid Itokawa for the rover to reach target position $r_f = [222.5 \quad -49.89 \quad 49.26]\ m$ from its initial position $r_0 = [-144 \quad 53.21 \quad 84.67]\ m$.

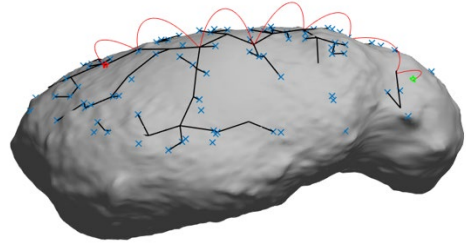

```
GENERATE_RANDOM_SAMPLE (r_init, r_goal, K, Δ)
    1   T.init(r_init)
    2   For k = 1 to K do
    3       r_rand ← RANDOM_STATE();
    4       r_near ← NEAREST_NEIGHBOR(r_rand, T);
    5       r_new ← STEER(r_rand, r_near, Δ)
    6       T.add_vertex(r_new);
    7       T.add_edge(r_near, r_new);
    8   end
    9   Return RANDOM_SAMPLE ← SHORTEST_PATH(r_init, r_goal, T)
```

**Figure 4: (Left) Algorithm for generating a random sample. (Right) A random sample created for the rover to reach a goal location from its initial location.**

The random sampler provides multiple feasible trajectories which are then optimized using Evolutionary Algorithms (EA). For a feasible trajectory from an initial location to a goal location with $n$ hops, the objective of the optimization problem is to minimize the sum of initial velocities for each hop. To avoid penetration of each hopping trajectory into the surface of the asteroid, the initial phase and final phase are constrained in two cones defined by angles $\theta_1$ and $\theta_2$ as shown in Figure 5 (left). Moreover, another constraint is added such that the initial velocity $v_{0(i)}$ for hop $i$ does not exceed the escape velocity around that location $v_{e(i)}$ as shown in Eq. (9).

$$\min_{\Gamma,\Pi,\Lambda} f(\Gamma, \Pi, \Lambda) = \sum_{i=1}^{n} v_{0(i)}$$

$$s.t. \quad c_1(\Gamma, \Pi, \Lambda) = v_{0(i)} - v_{e(i)} \leq 0$$

$$c_2(\Gamma, \Pi, \Lambda) = \theta_{1(i)} - 45° \leq 0$$

$$c_3(\Gamma, \Pi, \Lambda) = \theta_{2(i)} - 45° \leq 0 \qquad (9)$$

Where, $\Gamma = [\tau_1 \quad \tau_2 \cdots \tau_n]$ is an $(n \times 1)$ array of transfer time for each hop, $\Pi = [r_1 \quad r_2 \cdots r_{n+1}]$ is an $\{(n+1) \times 3\}$ array of the position of the robot for each hop such that the rover sequentially hops from position $r_i$ to $r_{i+1}$ and $\Lambda = [v_{0(1)} \quad v_{0(2)} \cdots v_{0(n)}]$ is an $(n \times 1)$ array of initial velocity for each hop. The cost function for the optimization problem is implemented by using the penalty method as shown in Eq. 10.

$$\min_{\Gamma,\Pi,\Lambda} J(\Gamma, \Pi, \Lambda) = f(\Gamma, \Pi, \Lambda) + 100 \sum_{i=1}^{3} g(c_i(\Gamma, \Pi, \Lambda))$$

$$where, \quad g(c_i(\Gamma, \Pi, \Lambda)) = \max(0, c_i(\Gamma, \Pi, \Lambda))^2 \qquad (10)$$

Figure 5 (right) shows the block diagram of the path planner using Evolutionary Algorithm (EA). $N$ random samples are created by varying the constraint $\Delta$ for each sample according to a uniform distribution, $\Delta \sim \mathcal{U}(a, b)$. The $N$ samples are taken as the initial population for the EA which undergoes successive crossover and mutation. During mutation, the variables $\Gamma$ and $\Pi$ are varied by a gaussian distribution as $\Gamma \sim \mathcal{N}(\Gamma, \sigma_\Gamma^2)$, $\Pi \sim \mathcal{N}(\Pi, \sigma_\Pi^2)$. The fitness of each individual is then calculated according to the cost function $J(\Gamma, \Pi, \Lambda)$, which are then ranked. The fittest $N$ individuals are then carried along to the next generation. Moreover, at each generation a new population is introduced to the EA by generating $N/2$ samples using the random sampler.

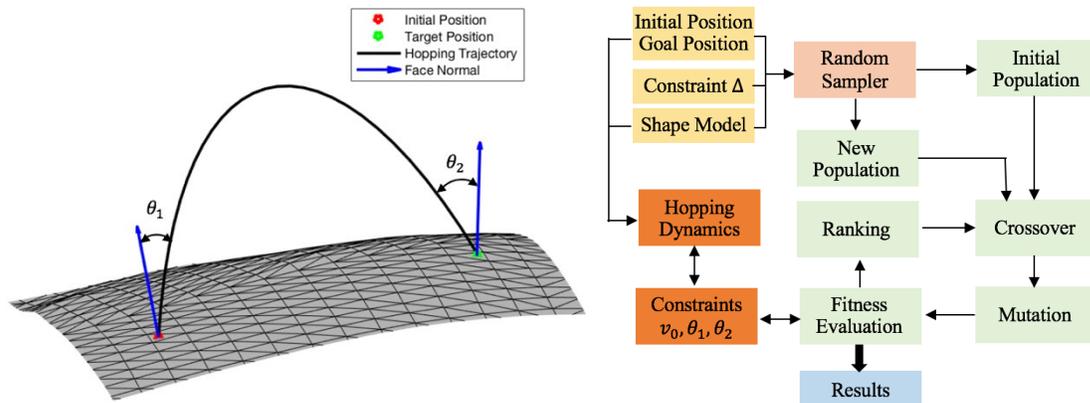

**Figure 5:** (Left) Constraints in $\theta_1$ and $\theta_2$ during the initial phase and the landing phase. (Right) Schematic for the optimized path planner using Evolutionary Algorithm.

Figure 6 (left) shows the optimized trajectory for a rover generated by the path planner to reach the target position $r_f = [222.5 \quad -49.89 \quad 49.26]m$ from its initial position $r_0 = [-144 \quad 53.21 \quad 84.67]m$. Figure 6 (right) shows the average cost function and cost function of the best individual over 51 generations. The bars show the standard deviation of each generation over 10 iterations. It can be seen that the standard deviation of the best individual reaches zero after 29 generations which shows that the path planner provides a solution with the same fitness at every iteration.

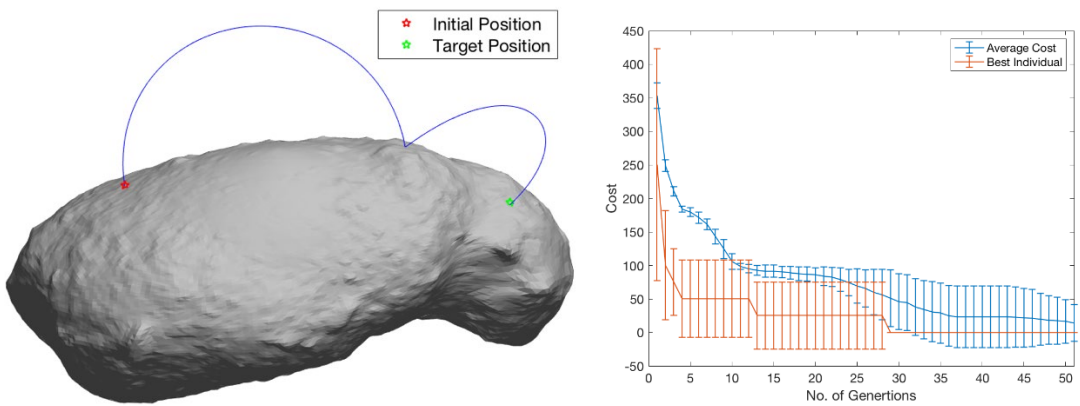

**Figure 6:** (Left) Optimized trajectory of a rover generated by the path planner to reach a target location from its initial location. (Right) Average cost function and cost function of the best individual over 51 generations. The bars show the standard deviation of each generation over 10 iterations.

Moreover, the constraint $\Delta$ can be fixed for generating the random samples depending on the hopping range of the rover. Figure 7 (left) shows the optimized trajectory for a rover to reach the target position $r_f = [222.5 \quad -49.89 \quad 49.26]\,m$ from its initial position $r_0 = [-144 \quad 53.21 \quad 84.67]\,m$ with $\Delta = 50\,m$. Figure 7 (right) shows a different scenario where the rover needs to reach the target position $r_f = \begin{bmatrix} 281.4 & -10.15 & 15.31 \end{bmatrix} m$ from its initial position $r_0 = \begin{bmatrix} -239.4 & 88.88 & 17.58 \end{bmatrix} m$ with two waypoints in between $w_1 = \begin{bmatrix} -64.27 & -122.3 & 40.22 \end{bmatrix} m$ and $w_2 = \begin{bmatrix} 135.9 & -40.59 & 107.9 \end{bmatrix} m$, and $\Delta= 50\ m$.

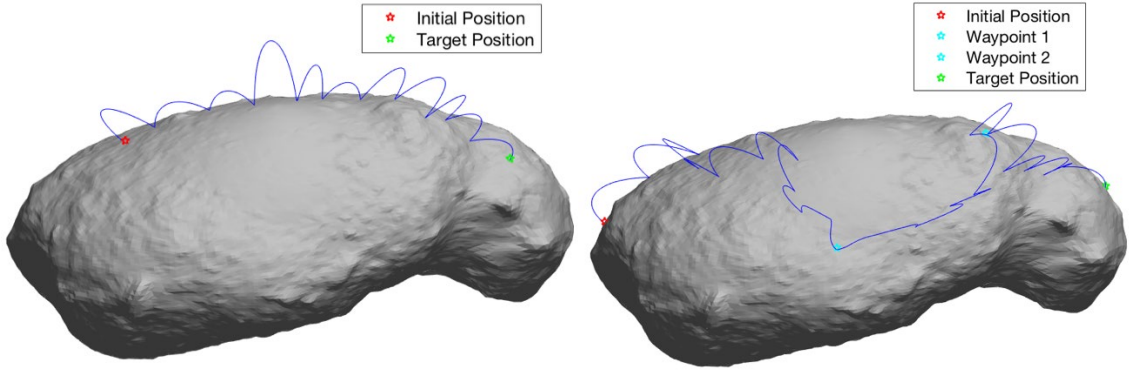

**Figure 7: (Left) Optimized trajectory for a rover to reach a goal location from its initial location with an $\Delta= 50m$ for each hop. (Right) Optimized trajectory for a rover to reach a goal location from its initial position with two waypoints and $\Delta= 50m$.**

## EXPLORATION WITH MULTIPLE ROVERS

Utilizing the motion planning strategies discussed above, we can enable a swarm of rovers to achieve mobility on the asteroid surface and perform exploration in parallel. The multiple rovers deployed on an asteroid surface are required to achieve maximum area coverage while maintaining multiple communication links so that acquired science data maybe communicated effectively back to a mothership. In this section, we describe an algorithm developed to distribute a fleet of $N$ rovers on an asteroid surface using the concept of virtual forces to repel each rover from each other and attracting them when a communication constraint is violated.[21] Each of the rovers have a sensing range of $r_s$ and a communication range of $r_c$ and they can communicate its location and orientation to its neighbors. By using the concept of virtual forces, we simulate the collective control of the rovers. The rovers interact with each other through a combination of global repulsion combined with local, limited attraction. Considering a network of $N$ rovers with position vector $r_i, i = 1,2,...,N$ and $\|r_{ij}\|$ representing the Euclidean distance between rover $i$ and $j$, the repulsion force is defined as Eq. 11.

$$f_r(i,j) = \frac{r_i - r_j}{\|r_{ij}\|^2} \tag{11}$$

The repulsive force $f_r$ causes the rovers to move away from each other to maximize exploration of a target area. Moreover, the attraction force is defined as Eq. 12.

$$f_a(i,j) = \begin{cases} r_i - r_j & if\ degree < D \\ 0 & otherwise \end{cases} \tag{12}$$

The attraction force $f_a$ constrains the degree of communication links for each rover by attracting rovers (locally) when they are on the verge of losing connection maintaining at least $D$ degrees. Finally, the net force experienced by rover $i$ is formulated as Eq. 13.

$$f(i) = \sum_{j=1, j\neq i}^{N} \left(f_r(i,j) + f_a(i,j)\right) \tag{13}$$

With the net virtual force experienced by rover $i$ calculated, the next hopping location is determined as, $r_i(next) = \alpha f(i)$, where, $\alpha$ is a proportionality constant dependent on the hopping capabilities of the rover. Figure 8 shows the position of each rover over iterations 1, 3, 5, 7, 9, 11, 13 and 15. The yellow cubes represent the rovers and the red lines connecting them shows the active communication links. 15 rovers were deployed at random positions on the surface of asteroid Itokawa and each rover has a communication range of $r_c = 100m$ and can hop a maximum distance of 30m. The rovers need to explore the maximum area while maintaining at least 2 communication links. It can be seen that the rovers were able to successfully move away from each other maintaining the desired communication links.

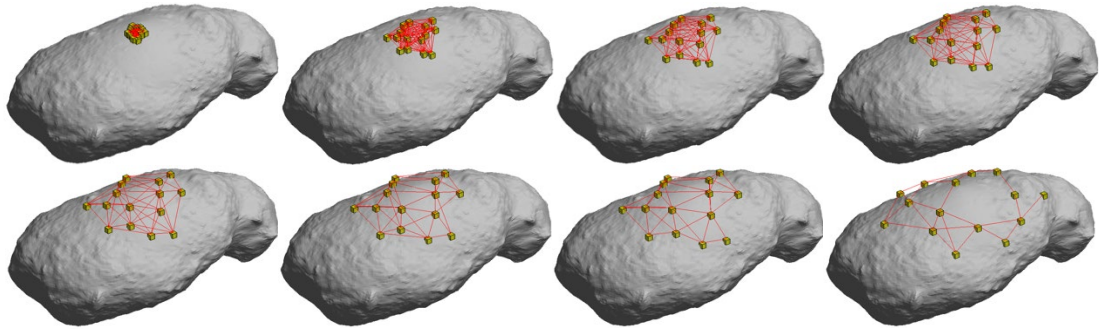

**Figure 8: Multiple rovers exploring the surface of asteroid Itokawa cooperatively. The yellow cubes represent the rovers and the red lines connecting them shows the active communication links.**

## CONCLUSION

In this paper, we presented a model-based path planning algorithm for hopping rovers on the surface of an asteroid with irregular gravity fields. By solving Lambert's boundary value problem in an irregular gravity field using a shooting method, hopping trajectories for a rover to impact a target location from its initial location is found. While in flight, the rover can localize itself by matching successive scans form a 3D LiDAR. The ballistic hopping dynamics and localization estimates are then used to develop an algorithm that provides optimal sequential hopping trajectories for a rover to reach a target position from its initial position over long distances. It is done by generating random samples constructing rapidly-exploring random trees on the asteroid surface constrained by maximum hopping capabilities of the rover. These feasible samples are then optimized further using genetic algorithms by successive crossover, mutation and fitness evaluation according to a cost function to generate near optimal trajectories to reach the desired goal. The algorithm can also drive the rover through multiple waypoints and eventually to the desired goal. Finally, an algorithm is demonstrated for multiple rovers to parallelly explore maximum area on an asteroid surface while maintaining a desired communication link among them. Our future work will include finding methods for multiple rovers to explore the entire surface of an asteroid using this path-planning algorithm.